\documentclass[11pt]{article}
\usepackage{graphicx}
\usepackage{wrapfig}
\usepackage{amsmath}
\usepackage{gensymb}
\usepackage{amsfonts}
\usepackage{multirow}
\usepackage{subfigure}
\usepackage{algorithm}
\usepackage{textcomp, gensymb}
\usepackage[nonatbib, preprint]{neurips_2022}
\usepackage[utf8]{inputenc} % allow utf-8 input
\usepackage[T1]{fontenc}    % use 8-bit T1 fonts
\usepackage{hyperref}       % hyperlinks
\usepackage{url}            % simple URL typesetting
\usepackage{booktabs}       % professional-quality tables
\usepackage{amsfonts}       % blackboard math symbols
\usepackage{nicefrac}       % compact symbols for 1/2, etc.
\usepackage{microtype}      % microtypography
\usepackage{xcolor}         % colors

\usepackage{algorithmicx, algpseudocode}
\algrenewcommand\algorithmicindent{1.0em}
\graphicspath{{Figures/}}
\usepackage{geometry}
\date{}
\newcommand{\repository}{\url{https://github.com/Zi-hao-Wei/Contrastive-Random-Walk-as-Intrinsic-Rewards}}
\title{Discovering Intrinsic Reward with Contrastive Random Walk}

\author{
  Zixuan Pan\\
  Department of Computer Science Engineering\\
  University of Michigan\\
  Ann Arbor, USA \\
  \texttt{zxp@umich.edu}
  \And
  Zihao Wei\thanks{Corresponding author: Zihao Wei, zihaowei@umich.edu}\\
  Department of Computer Science Engineering\\
  University of Michigan\\
  Ann Arbor, USA \\
  \texttt{zihaowei@umich.edu} \\
  \And
  Yidong Huang\\
  Department of Computer Science Engineering\\
  University of Michigan\\
  Ann Arbor, USA \\
  \texttt{owenhji@umich.edu} \\
  \And
  Aditya Gupta\\
  Department of Electrical and Computer Engineering\\
  University of Michigan\\
  Ann Arbor, USA \\
  \texttt{adigupta@umich.edu} \\
  \\
  \repository
}

\begin{document}
\maketitle

\begin{abstract}
    \noindent The aim of this paper is to demonstrate the efficacy of using Contrastive Random Walk as a curiosity method to achieve faster convergence to the optimal policy. Contrastive Random Walk defines the transition matrix of a random walk with the help of neural networks. It learns a meaningful state representation with a closed loop. The loss of Contrastive Random Walk serves as an intrinsic reward and is added to the environment reward.  Our method works well in non-tabular sparse reward scenarios, in the sense that our method receives the highest reward within the same iterations compared to other methods. Meanwhile, Contrastive Random walk is more robust. The performance doesn't change much with different random initialization of environments. We also find that adaptive restart and appropriate temperature are crucial to the performance of Contrastive Random Walk.
\end{abstract}

\section{Introduction}
Reinforcement learning (RL) learns policies that maximize the reward given by the environment. Methods considering only the environment rewards work well when rewards are dense. However, when the rewards are sparse, exploration has been proven important to make faster convergence of finding optimal policy. In the past few decades, dozens of methods have been proposed to encourage exploration by introducing an exploration bonus as an intrinsic rewards, which decides based on the chosen time of a certain state\cite{bellemare2016unifying,auer2002using,ICM}. 

With the advance of RL, people are interested in applying it to more complicated scenarios, where the observations are some high-dimensional signals (e.g. video frame)\cite{mnih2015human}. To understand the signals and better define states, instead of human definition, features extracted by neural networks are applied to represent states\cite{mnih2015human,info_gain}. It provokes two notable changes in the definition of states: states become high-dimensional and states are no longer invariant to observations. These changes make counting the visits to certain states unavailable.

To deal with the problem that traditional exploration rewards can't be used in complicated environments, researchers propose the concept of curiosity and use neural network loss as rewards\cite{ICM}. Neural networks are good at understanding high-dimensional signals and have astonishing generalization ability\cite{neyshabur2018towards}, which perfectly fits the scenario. The loss of neural networks can be viewed as a pseudo-count, that is lower when the state is frequently visited and higher for un-visited states. 

However, two problems remain with many existing curiosity methods. On one hand, many of the methods lack mathematical foundations. The intuition of why these methods work is unclear. On the other hand, neural networks in most curiosity methods neural networks only serve as function fitters and are not extracting meaningful features. We argue that these neural networks can also help build a better state encoder if properly designed. Good state representations can thus further leverage the generalization ability of neural networks and implicitly create a good clustering of observations. Such clusters can well define curiosity for similar observations, i.e. intrinsic rewards will not decrease when seeing exactly the same observations, but will also decrease for similar states. More meaningful representations will be better at defining what is "similar".

To tackle these two problems, we proposed applying Contrastive Random Walk (CRW) as a curiosity method\cite{jabri2020space}. CRW can extract meaningful representations from time series by creating a closed-loop for the series. In RL, experience is naturally a time series and hence CRW could also be able to extract good state representations. Meanwhile, we provide a mathematical proof that the loss of CRW is a good approximation to the information gain, which is a good measurement of how visited a state is.

We tested our method on two environments: "MountainCar-v0" and "Montezuma's Revenge". Both environments have sparse environment rewards and thus good exploration strategies are critical to ensure a faster convergence of policy. Meanwhile, both environments use neural networks to get state representations. MountainCar-v0 is a simple environment where the agent's task is to reach the top of the mountain within less time. The observation is some attributes of the car, including speed and position. While Montezuma's Revenge is a more complicated environment... The observation is just video frames. Our method has obtained better performance than all the other baselines, including the curiosity method, count-based method, and randomized strategy. Meanwhile, our methods are less subject to randomness of environments and randomness of policy that can give similar performance on all kinds of random initialization compared to other methods.

%TODO: Add environment specifics

\section{Preliminary and Notations}

\subsection{Reinforcement Learning}
Given a Markov decision process (MDP), which can be described as $\mathcal{M}=\langle\mathcal{S}, \mathcal{A}, P, E, \gamma\rangle$, where $\mathcal{S}$ is the state space, $\mathcal{A}$ is the action space, $P\left(s_{t+1} \mid s_{t}, a_{t}\right)$ is the probability distribution over transitions, $E\left(s_{t}, a_{t}\right)$ is the environment reward function, and $\gamma$ is the discount factor. We will focus on infinite-horizon MDPs.

\subsection{Intrinsic Reward}
We define an exploration bonus based on the current state $I(s_{t})$ and add it to the reward $E\left(s_{t}, a_{t}\right)$ given by the environment, getting a total reward  $R\left(s_{t}, a_{t}\right)=E\left(s_{t}, a_{t}\right)+I(s_{t})$.  The goal of the addition of exploration bonus is to embed the new reward to existing reinforcement learning algorithms and help to make a better policy.

\subsection{Non-tabular State Representation}
The non-tabular environment of reinforcement learning can be viewed as an infinite-horizon stochastic dynamic programming, where states are some high-dimensional representations extracted by neural networks, denoted as $s = \phi(o)$, $s = [s_1, s_2, ..., s_n]^T \in \mathbb{R}^n$, where $o$ is the observation (video frame). A prominent characteristic of non-tabular states is that the state representations are variant. As a neural network is applied and updated with experience, state representation of observations will change (i.e. $s$ will be updated with time). performance for count-based methods. 

% TODO: Add bellman equation for non-tabular cases

\section{Related Works}

\subsection{Curiosity Driven Intrinsic Reward}
In order to make count-based methods also work for non-tabular state representations, discrete "counts" are changed into some density and continuous functions\cite{bellemare2016unifying}. However, this approximation also incurs performance drop for count based methods. Therefore, curiosity methods are proposed to create a new kind of exploration bonus for non-tabular states.

Curiosity driven intrinsic rewards encourages the agent to explore the environment by learning an intrinsic reward $I(s_{t})$ besides the environment reward\cite{RND,ICM,burda2018large,disagreement,de2018curiosity}. The intrinsic reward is given by neural network loss. $$ I(s_{t}) = L_{nn}(s_{t}) + C $$
where C is a constant.

It applies the fact that if an input has been seen, loss of the neural network will decrease, indicating a lower intrinsic reward. Neural network loss style intrinsic reward also applies the generalization ability that makes it suitable for continuous state representation\cite{neyshabur2018towards}. The loss will drop for exactly the same input $s_t$, but also states with similar feature to $s_t$, which makes it appropriate for non-tabular tasks. 

There are mainly two types of curiosity methods: action dependent methods and state-only methods. State-only methods only use state representations from observations as inputs to the neural network\cite{RND,disagreement,mohamed2015variational}. Action dependent methods take both actions and state representations as input into neural networks\cite{ICM,de2018curiosity}.

\subsection{Information gain as Measurement of Intrinsic Reward}
The information gain or mutual information is defined as:
$$
IG(X ; Y)=\mathbb{E}_{p(x, y)} \log \frac{p(x, y)}{p(x) p(y)}=H(X)-H(X \mid Y)
$$

It is widely used in Reinforcement learning as a measurement of intrinsic reward\cite{bellemare2016unifying,de2018curiosity}.

For curiosity methods, information gain is made larger for unvisited states, while lower for frequently visited states to encourage the agent exploring the environment.

There exist various definitions of information gain in Reinforcement Learning\cite{info_gain,tiomkin2017unified}. We will use the state-only transition information, which has been proved to be a good indicator for state representation\cite{info_gain}.

$$IG(s_{t+k};s_{t}) = H(s_{t+k}) - H(s_{t+k}|s_{t})$$

In non-tabular scenario, we can change it to $$IG(s_{t+k};s_{t}) =\Sigma_{i=1}^n H(s_{t+k, i}) - H(s_{t+k, i}|s_{t, i})$$. And the intrinsic reward for $s_{t+k}$ is thus 
$$I(s_{t+k}) = IG(s_{t+k};s_{t})$$

\subsection{State Representation Learning}
There are a bunch of previous works on how to extract better state representations\cite{stooke2021decoupling,nachum2018near,info_gain,hjelm2018learning}. Better representations have been proved to be crucial to make RL algorithms perform better in many scenarios. Our work is most similar to \cite{hjelm2018learning}, where the information gain is used to evaluate how good a representation is. The goal of this paper is to maximize information gain between inputs and outputs of the state encoder. Our method is also arguably increasing such information gain.

\section{Contrastive Random Walk}

\begin{figure*}[htbp]
    \centering
    \includegraphics[scale=0.4]{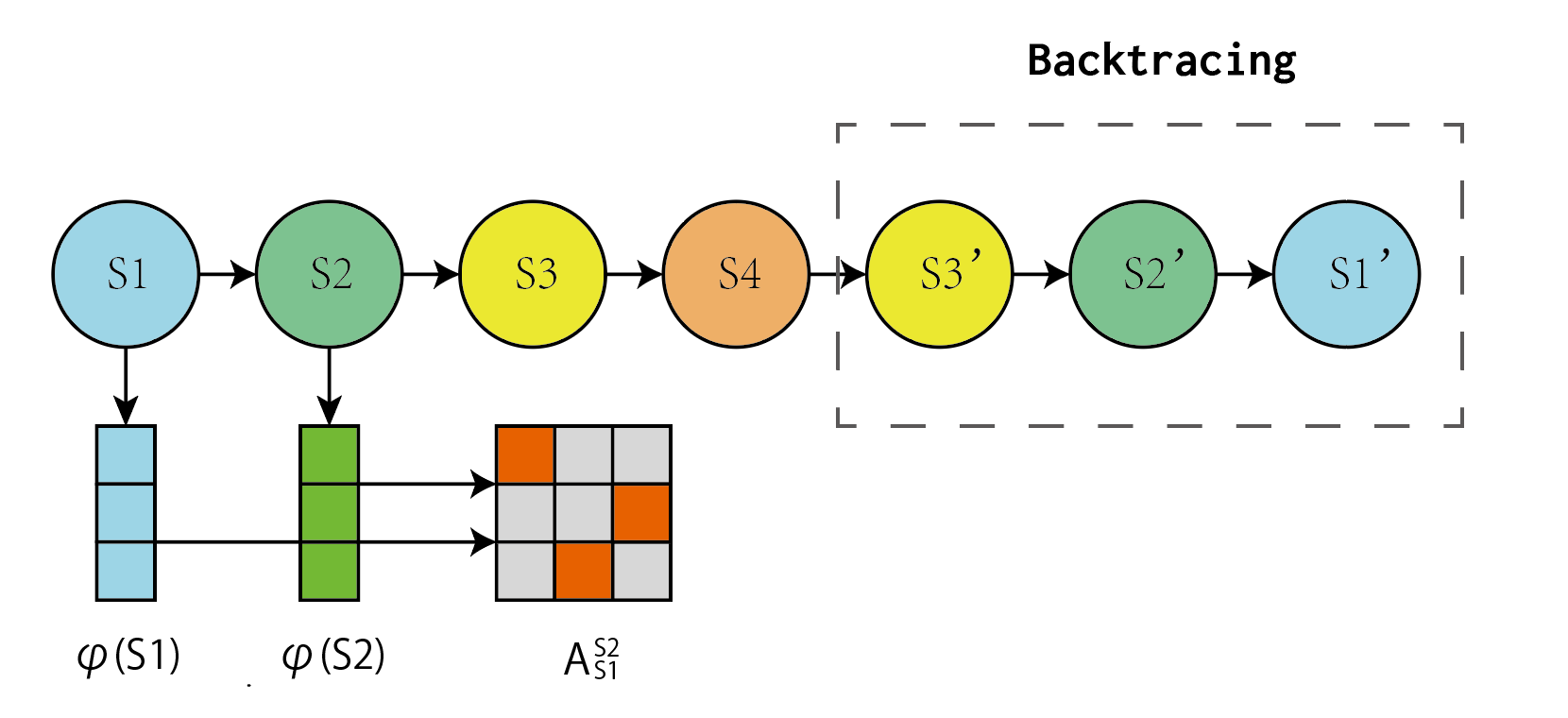}
    \caption{Overview of Contrastive Random Walk. A backtracing process is appended to the forward path. States are connected with affinity matrices.}
    \label{fig:model}
\end{figure*}

Inspired by previous work, we view each experience as a reversible process\cite{jabri2020space}. Given an experience starting with $s_1$, ending with $s_t$, we also obtained another experience starting with $s_t$, ending with $s_1$. We concatenate the reverse experience to the original experience as shown in Figure\ref{fig:model}. We give each state an embedding, denoting the encoder as $\phi$.
$$A_{t}^{t+1}(i, j)=\operatorname{softmax}\left(f(s_{t} s_{t+1}^{\top})\right)_{i j}=\frac{\exp \left(d_{}\left(\mathbf{s}_{t}^{i}, \mathbf{s}_{t+1}^{j}\right) / \tau\right)}{\sum_{l=1}^{N} \exp \left(d_{}\left(\mathbf{s}_{t}^{i}, \mathbf{s}_{t+1}^{l}\right) / \tau\right)}$$
where $d_{}\left(s_{1}, s_{2}\right)=\left\langle\left(s_{1}\right), \left(s_{2}\right)\right\rangle$ is the similarity between embeddings. $\tau$ is the temperature, which is a constant in softmax. By using the softmax, we make this affinity matrix also a transition matrix. Then we can define a continuous transition from t to t+k.

$$\bar{A}_{t}^{t+k}=\prod_{i=0}^{k-1} A_{t+i}^{t+i+1}=P\left(s_{t+k} \mid s_{t}\right)$$
Intuitively, it just predicts the state of t+k given state of t. Our goal is defined as:
$$\mathcal{L}_{c y c}^{k}=\mathcal{L}_{C E}\left(\bar{A}_{t}^{t+k} \bar{A}_{t+k}^{t}, I\right)=-\sum_{i=1}^{N} \log P\left(s_{t+2 k, i} \mid s_{t, i}\right)$$
When the objective function is minimized, each affinity matrix will be a one-hot matrix containing only 0 and 1, with one 1 in each row and column. This is to give a one-by one map from previous state the current state. Each node in the previous state embedding aims to maximize the similarity with another node in the current state, and minimize the similarity with all the other nodes in the current state, which can be viewed as a contrastive learning with n-1 negative samples. 

For the intrinsic reward at each state, we use 
$$I(s_{t}) = \mathcal{L}_{c y c}^{k} + C$$ starting from state $s_t$ and ends with $s_{t+k}$. A simple choice of $k$ will be 1 for all states if we use the replay buffer and Deep Deterministic Policy Gradient algorithm. If the batch contains the entire experience, we can set $k > 1$. It can provide implicitly supervision to intermediate states.

\subsection{Adaptive Restart}
Since the environment we use is quite simple, the convergence of CRW is often too fast, and results in the intrinsic reward approaching zero with more iterations. Therefore, we applied the strategy of adaptive restart in optimization to settle the problem\cite{o2015adaptive}. The parameters of CRW encoder are reset after the average intrinsic reward is smaller than a threshold. A psuedo code of our method can be found in Algorithm \ref{alg:forward}.

\begin{algorithm}
\caption{model.forward()}
\label{alg:forward}
\hspace*{\algorithmicindent}\textbf{Input:} $x$, threshold, temperature

\hspace*{\algorithmicindent} \#x size: $B\times T\times N$
\begin{algorithmic}
\State A = einsum("bti,btj$\rightarrow$ btij" x[:,:,:-1], x[:,:,1:]) / temperature
\State \#affinity matrix of forward path with size $B\times T\times N\times N$
\State A\_ = A.transpose(-2,-1)
\State for i in range(T):
\State \quad AA[i] = bmm(A[:,i],A\_[:,i])
\State AA = multiply(AA)
\State \# Multiply all elements in AA
\State loss = sum(-log(diag(AA,dim=-2,-1)))
\State \# Compute Cross Entropy loss.
\State If loss $\leq$ threshold:
\State \quad reset\_param()
\State \# Adaptive restart
\end{algorithmic}
\end{algorithm}

\subsection{Approximation to Information Gain}
For the contrastive random walk,
\begin{gather*}
    IG(s_{t+2k};s_{t}) =\Sigma_{i=1}^n H(s_{t+k, i}) - H(s_{t+2k, i}|s_{t, i})\\
    =\Sigma_{i=1}^n \mathbb{E}_{s_{t+2k}\sim \rho^{\prime}}[log(s_{t+2k,i})] + \mathbb{E}_{p(s_{t+2k})\sim \rho^{\prime},s_{t}\sim \rho}[log(p(s_{t+2k,i}|s_{t,i}))]
\end{gather*}
where $\rho^{\prime}$ is the discounted state distribution for backtracing states, and $\rho$ is the discounted state distribution for forward states.

Because the backtracing gives exact the same states of forward path, $$\mathbb{E}_{s_{t+2k}\sim \rho^{\prime}}[log(p(s_{t+2k,i}))] = \mathbb{E}_{s_{t+2k}\sim \rho}[log(p(s_{t,i}))]$$ is not determined by the contrastive random walk, we just replace it by a constant $\mathrm{C}$. 

Contrastive random walk optimizes the second term, such that:
    $$\mathbb{E}_{p(s_{t+2k})\sim \rho^{\prime},s_t\sim \rho}[log(p(s_{t+2k,i}|s_{t,i}))] = \frac{\Sigma_{m=1}^Mp_m(s_{t+2k,i}|s_{t,i})log(p_m(s_{t+2k,i}|s_{t,i}))}{M}$$ 
Denote $P = \max p_m$, and when $P$ is close to 1, the above equation can be approximated by:
$$\mathbb{E}_{p(s_{t+2k})\sim \rho^{\prime},s_t\sim \rho}[log(p(s_{t+2k,i}|s_{t,i}))] \approx \frac{log(P(s_{t+2k,i}|s_{t,i}))}{M} $$
Remove the constant $M$, we get
$$\mathbb{E}_{p(s_{t+2k})\sim \rho^{\prime},s_t\sim \rho}[log(p(s_{t+2k,i}|s_{t,i}))] \approx log(P(s_{t+2k,i}|s_{t,i})) $$
This approximation is sufficient as the goal of contrastive random walk is to maximize $P$.

Thus, we can rewrite the information gain as:
\begin{gather*}
    IG(s_{t+2k};s_{t}) = \mathbb{E}_{s_{t+2k}\sim \rho}[log(p(s_{t,i}))] + L_{cyc}^k
\end{gather*}

Then, by optimizing the information gain of $s_{t+2k}$, we are also optimizing the information gain of $s_{t}$, as they share the same state representation.

\begin{gather*}
    I(s_{t}) = I(s_{t+2k}) = IG(s_{t+2k};s_{t})
\end{gather*}

\section{Experiments}
\subsection{Training Settings}
\subsubsection{Environments}
\begin{figure*}[htbp]
    \centering
    \includegraphics[scale=0.5]{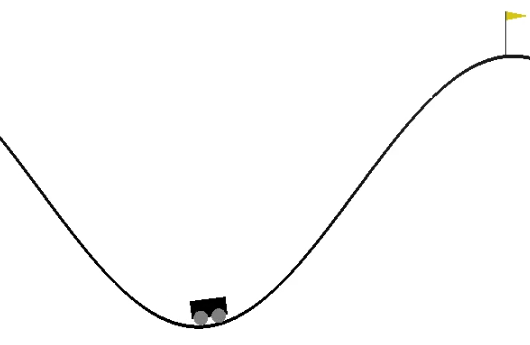}
    \caption{Overview of the MountainCar-V0 environment.}
    \label{fig:MountainCar}
\end{figure*}

Because of the lack of computational resources, we use a simple environment called "MountainCar-v0" \cite{Moore90efficientmemory-based}, to test our methods and other baselines. In the MountainCar-v0 environments(shown in Figure \ref{fig:MountainCar}), a car is originally located in the valley and the agents are forced to learn a policy which can drive up the car from the bottom to the mountain on the right. However, since the car's engine is not strong enough, the only way to succeed is to drive back and forth to build up momentum. The entire environment has an observation of 2 dimension $(o_1,o_2)$, denoting the position of the car on the X-axis and the velocity of the car; three action space $(a_1,a_2,a_3)$, standing for accelerating right, left, or not. In order to encourage the agents to arrive at the destination as fast as possible, a reward $r$ of -1 is given at each step when the car hasn't arrived, and the total steps in one episode are 200, meaning the reward is ranging from -200 to 0. 

\subsection{Implementation Details}
To test the effectiveness of the exploration bonus we proposed, we combine our CRW Intrisic Reward with a Deep Q Network(DQN) backbone. Here we use a three-layer MLP with the default hidden dimensions (256,128,64) with 0.3 dropout ratio as our state encoder $\phi$ and Q-network. We compare our methods with several exploring baselines, which are Epsilon-greedy, Upper Confidence Bound (UCB), and Random Network Distillation(RND). These baseline models will use the same DQN as backbone. Q value is updated with the function
$$
\tilde{Q}(s_t, a_t) \longleftarrow\left(1-\beta_{k}\right) \tilde{Q}(s_t, a_t)+\beta_{k}\left(E(s_t, a_t)+I(s_t)+\alpha \max _{v} \tilde{Q}(s', v)\right)
$$
The target and predictor network of RND will share the same architecture of state encoder of our CRW. The main hyper-parameters and their default value used in the experiments are, temperature $\tau=1$, DQN learning rate $lr_1=1e-3$, state encoder learning rate $lr_2=1e-5$, restart running loss threshold $s=5e-3$, batch size $B=128$, replay buffer size $M=100000$, episode num $N=1000$. And the entire experiment is carried out on one NVIDIA RTX 3080 GPU. More details can be found in the code.

\subsection{Results}
\begin{figure*}[!htb]
	\centering
	\subfigure[Epsilon Greedy]{
		\begin{minipage}[b]{0.4\textwidth}
			\includegraphics[width=1\textwidth]{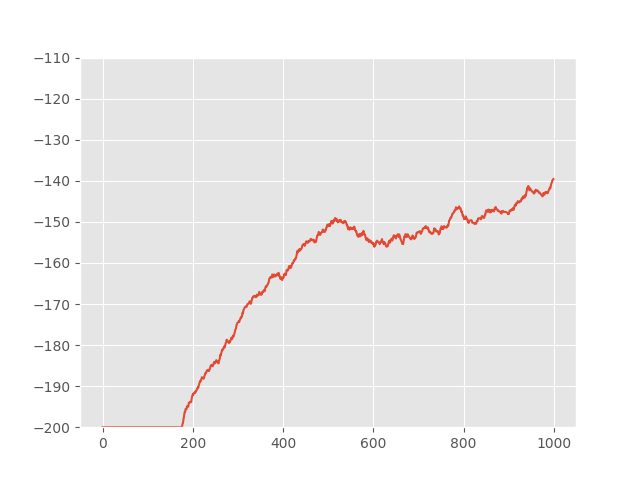} 
		\end{minipage}
		\label{fig:Greedy}
	}
	\subfigure[Upper Confidence Bound]{
		\begin{minipage}[b]{0.4\textwidth}
   	 	\includegraphics[width=1\textwidth]{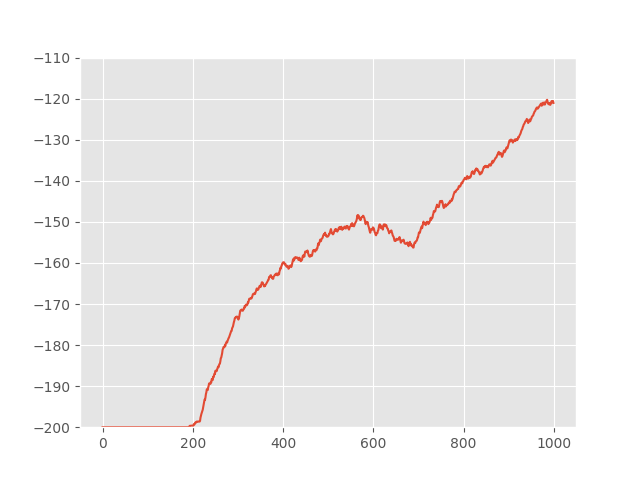}
		\end{minipage}
	\label{fig:UCB}
	}
	\\ 
	\subfigure[Random Network Distillation]{
		\begin{minipage}[b]{0.4\textwidth}
			\includegraphics[width=1\textwidth]{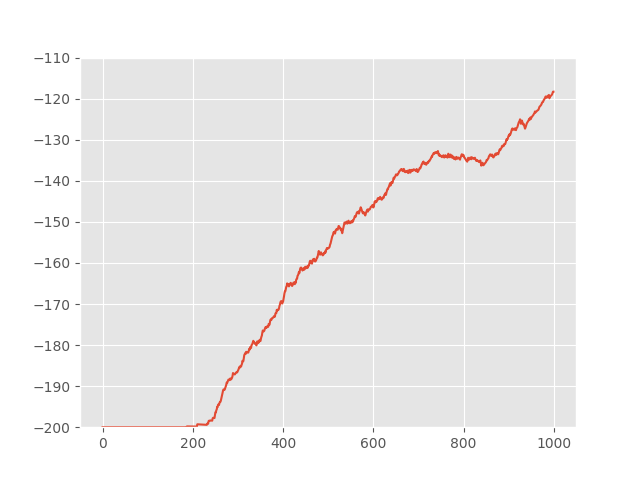} 
		\end{minipage}
		\label{fig:RND}
	}
	\subfigure[Contrastive Random Walk]{
		\begin{minipage}[b]{0.4\textwidth}
	 	\includegraphics[width=1\textwidth]{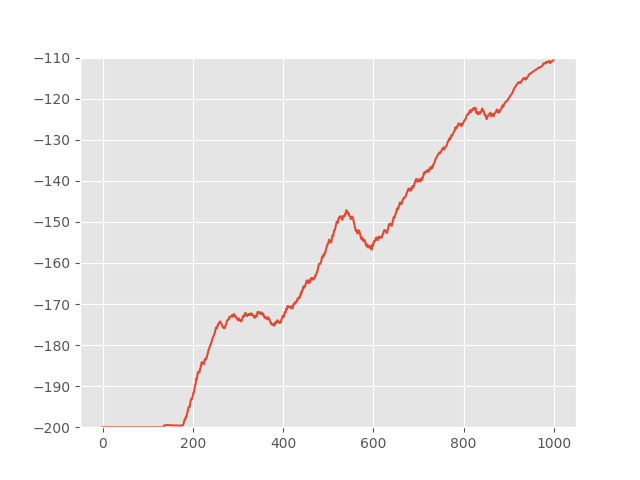}
		\end{minipage}
	\label{fig:CRW}
	}
	\caption{The running episode reward of the training process of CRW and different baselines.}
	\label{fig:Performance}
\end{figure*}

Here we report the performance of our CRW and other baselines in Figure \ref{fig:Performance}. Our methods reach a final episode running reward of -110.607, which greatly surpasses all the other baselines, where epsilon greedy achieves -139.57, UCB achieves -121.037 and RND achieves -118.297. Meanwhile, we also observe that our methods begin to have rewards earlier than other networks. This phenomenon shows that our method wastes less efforts on testing a similar states, which benefits the convergence speed.

\subsection{Ablation Study}
\begin{figure*}[!htb]
	\centering
	\subfigure[Tuning feature dimension in state encoder]{
		\begin{minipage}[b]{0.4\textwidth}
			\includegraphics[width=1\textwidth]{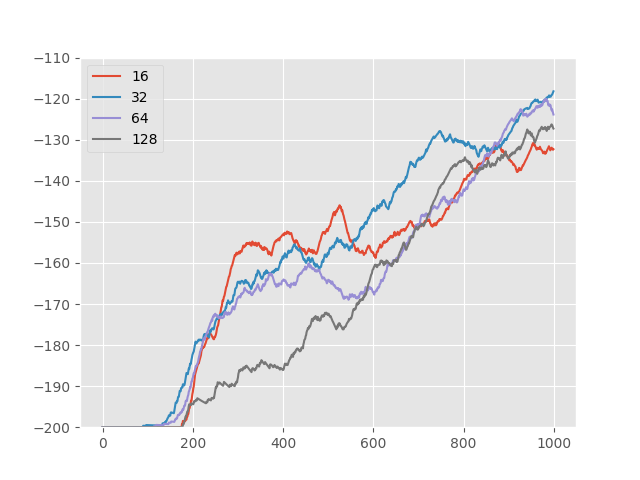} 
		\end{minipage}
		\label{fig:CFS}
	}
	\subfigure[Tuning crw temperature]{
		\begin{minipage}[b]{0.4\textwidth}
   	 	\includegraphics[width=1\textwidth]{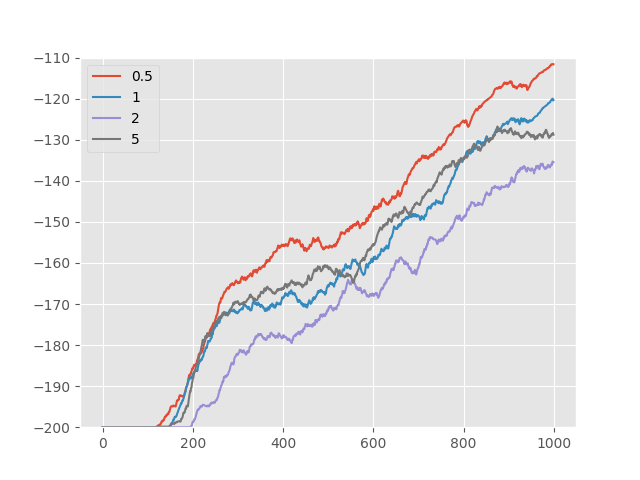}
		\end{minipage}
	\label{fig:CT}
	}
	\\ 
	\subfigure[Tuning learning rate of the state encoder]{
		\begin{minipage}[b]{0.4\textwidth}
			\includegraphics[width=1\textwidth]{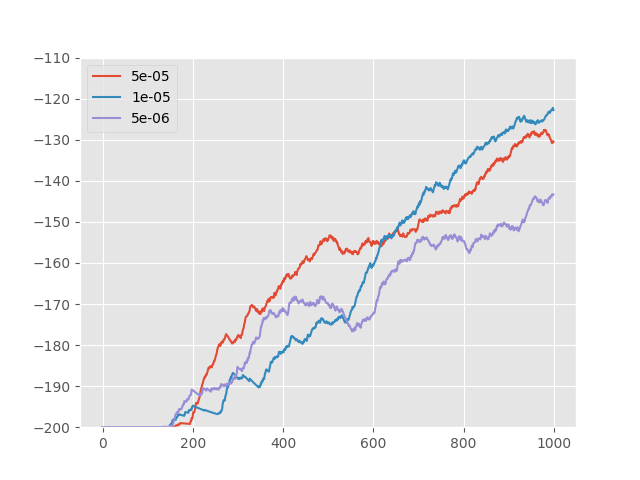} 
		\end{minipage}
		\label{fig:CSE}
	}
	\subfigure[Tuning restart running loss ratio]{
		\begin{minipage}[b]{0.4\textwidth}
	 	\includegraphics[width=1\textwidth]{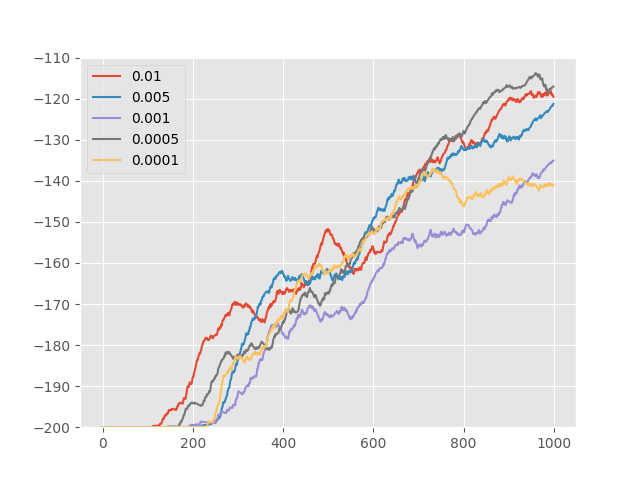}
		\end{minipage}
	\label{fig:CRL}
	}
	\caption{The running episode reward of the training process of CRW and different baselines.}
	\label{fig:Ablation}
\end{figure*}

Here we show how each hyperparameter will contribute to the final results of our methods in Figure \ref{fig:Ablation}. The hyper-parameter tuned including feature dimension of the state encoder, temperature of crw, batch size during training and learning rate. Except for the parameter of being tuned, other parameters are set as default.

\subsection{Discussion}
As shown in the results, temperature is an important factor to improve model performance. We find that the reward at 1000 iteration is linearly related to the choice of temperature. Generally, a small temperature with adaptive restart works better. Also, we found that adaptive restart is essential for CRW. When the restart threshold is set smaller, the performance of CRW may drop significantly.

\section{Experiments on more complicated environments}
\begin{figure*}[htbp]
    \centering
    \includegraphics[scale=1]{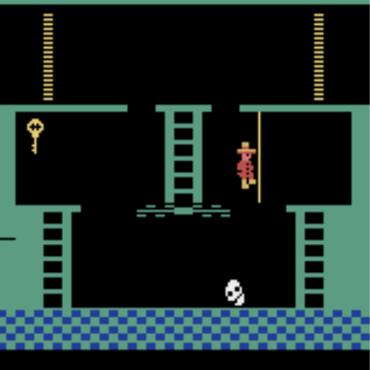}
    \caption{Overview of the MontezumaRevenge-v0 environment.}
    \label{fig:Montezuma}
\end{figure*}

Besides the  ``MountainCar-v0" \cite{Moore90efficientmemory-based} environment, we also tried to implement our algorithm on the classic atari game ``MontezumaRevenge-v0" \cite{ale} which has a sparse reward. As a traditional arcade game \ref{fig:Montezuma}, the observation space of the game is a RGB image with hieght 210 pixels and width 160 pixels. The aim for the agent is to go through different rooms and get as many points as possible. Former work \cite{RND} showed that with the help of intrinsic rewards, the agent can get out of the first room and gain points while normal RL algorithms  fail to get a single point. 

\subsection{Implementation Details}
For the agent network, we used an actor-critic architecture with cnn state encoder. We use a PPO \cite{ppo} training method to update the networks, which is a same architecture  suggested in the RND model \cite{RND}. To get the experience, we implement 64 workers using python multiprocessing. We also set the PPO learning rate $lr=1e-5$. For the CRW model that we proposed, we used a slightly different mean of encoding.

\noindent The pre-processing of the observations starts with  compressing the image into a 160*160 large image. Getting the compressed image, we split the image into 16 patches, each of 64*64 large. The overlapping region between neighboring patches is used to avoid missing information near the edges. After getting the 16 patches of images, we use a 4-layer convolution network with leakyrelu as activation layer to get the embedding of each patch. With these 16 embedding,  we implement the CRW model and get the intrinsic reward. To further extract property of the moving characters, we use a localized method, which also extract 16 smaller patches around the moving character to produce intrinsic reward.  We trained the network on a NVIDIA RTX 3080 GPU. More implementation details can be found in the code.

\subsection{Results and qualitative analysis}
\begin{figure*}[htbp]
    \centering
    \includegraphics[scale=0.7]{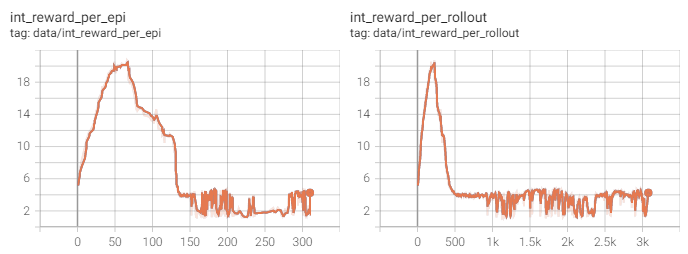}
    \caption{Result of implementation in MentonzumaRevenge-v0.}
    \label{fig:Montresult}
\end{figure*}

Because of the lack of computational resource, we just tried the implementation for 300 episodes and investigated the change of intrinsic reward. Because 300 episodes were not enough for the agent to get out of the first room, the implementation was in fact a purely exploration problem. As is shown in Figure \ref{fig:Montresult}, we can see that as the episode number increase, the intrinsic reward given by our CRW model first increase and then decrease, showing that the exploring first arise curiosity of the world, and then after many states are visited, the model lose the curiosity, thus the loss will drop to a lower value. This implies that our model can do the job of helping the algorithm to explore more unvisited states.

\section{Conclusion}
We demonstrated a new intrinsic reward with Contrastive Random Walk that achieves better performance than other methods on sparse-reward non-tabular scenarios. This method has a more solid mathematical foundation related to information gain, and whose neural network learns meaningful representations of states. These two advantages ensures our method also has the potential to provide good results in more complicated environments. Our success also provokes thinking on the importance of learning a good representation win the meantime of exploring the environment. It also encourages us to explore the mathematical proof of a faster conversion using this method as future work.

\bibliographystyle{plain}
\bibliography{citation}

\appendix
 
\end{document}